\documentclass[10pt, a4paper]{article}

\usepackage[final]{style} % this is the new style
\title{Advancing Academic Chatbots: Evaluation of Non-Traditional Outputs}

\name{Nicole Favero, Francesca Salute, Daniel Hardt} 

\address{Copenhagen Business School \\
         Copenhagen, Denmark \\
         nifa23ac@student.cbs.dk, frsa23ac@student.cbs.dk, dha.msc@cbs.dk\\
         }

\abstract{
Most evaluations of large language models (LLMs) focus on standard tasks such as factual question answering (Q\&A) or short summarization. This research expands that scope in two directions: first, by comparing two retrieval strategies, Graph RAG (structured knowledge-graph-based) and Advanced RAG (hybrid keyword-semantic search), for Q\&A; and second, by evaluating whether LLMs can generate high-quality non-traditional academic outputs, specifically slide decks and podcast scripts.
We implemented a prototype combining Meta’s LLaMA-3-70B (open-weight) and OpenAI’s GPT-4o-mini (API-based). Q\&A performance was evaluated using both human ratings across eleven quality dimensions and large-language-model (LLM) judges for scalable cross-validation. GPT-4o-mini with Advanced RAG produced the most accurate responses. Graph RAG offered limited improvements and led to more hallucinations, partly due to its structural complexity and manual setup.
Slide and podcast generation was tested with document-grounded retrieval. GPT-4o-mini again performed best, though LLaMA-3 showed promise in narrative coherence. Human reviewers were crucial for detecting layout and stylistic flaws, highlighting the need for combined human–LLM evaluation in assessing emerging academic outputs.
 \\ \newline }

\begin{document}

\maketitleabstract

\section{Introduction} 
Large Language Models (LLMs) are increasingly used in higher education to support content creation, advising, and research-driven Q\&A. Yet most evaluations still focus on standard tasks (factual Q\&A, short summaries) and overlook non-traditional, pedagogically oriented outputs such as slide decks and podcast scripts. This research broadens the scope by (i) comparing two retrieval strategies: Advanced RAG (hybrid lexical-semantic) and Graph RAG (structured knowledge graph), for academic Q\&A, and (ii) assessing whether LLMs can reliably generate document-grounded slides and podcasts suitable for instructional use.

\begin{figure}[!ht]
    \centering
    \includegraphics[width=1\linewidth]{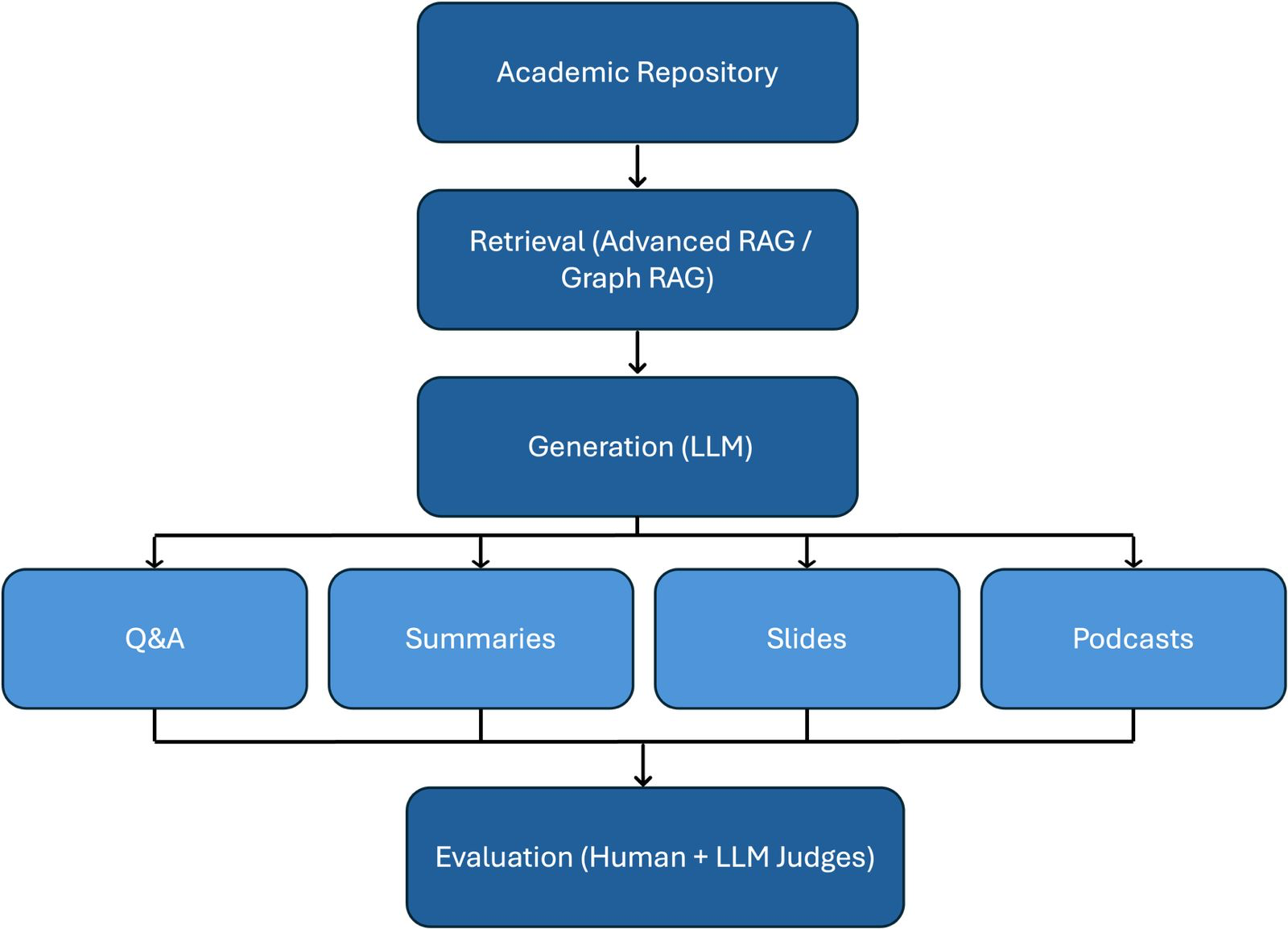}
    \caption{Framework overview for generating diverse academic outputs from a research repository.}
    \label{fig:framework}
\end{figure}

We implement a prototype combining an open-weight model (LLaMA 3.3 70B Instruct) and an API model (GPT 4o mini), each integrated with both retrieval pipelines. A dual-track evaluation, human raters and independent LLM-as-a-Judge systems, examines coherence, correctness, usefulness, hallucinations, and overall satisfaction across Q\&A, slides, and podcasts. Figure~\ref{fig:framework} presents the framework overview and dashboard flow: an academic repository feeds retrieval (Advanced RAG / Graph RAG), generation (LLM), and output modules (Q\&A, summaries, slides, podcasts), followed by combined human and LLM evaluation.

We address two questions: 
\begin{itemize}
    \item \textbf{RQ1}: Which retrieval pipeline best supports accurate, useful academic Q\&A across models?
    \item \textbf{RQ2}: Can LLMs, when grounded in source documents, produce high-quality slides and podcasts that meet pedagogical standards, and how closely do human and LLM evaluators align?
\end{itemize}
Our contributions are: a  comparison of Graph RAG vs. Advanced RAG in academic Q\&A, an assessment of non-traditional outputs for teaching, and a practical evaluation protocol that combines human nuance with scalable automated judgment.

\section{Literature Review}
\label{sec:literature}
This study is grounded in two primary areas of literature: the application of Large Language Models (LLMs) in educational contexts, particularly through chatbots, and the evaluation of LLM outputs, especially non-traditional educational artifacts like slide decks and podcasts.

\subsection{LLMs in Education}
LLMs are reshaping higher education by offering personalized feedback, real-time assistance, and adaptive content delivery. Chatbots provide continuous support to students, enabling learning across schedules and geographies \citep{studentchatbot, yigci2024llm}. For educators, they support course design, content creation, and task automation \citep{lopez2025llms}. Platforms such as \textit{Advisely} illustrate how LLMs can enhance academic advising by alleviating workload and improving information accessibility \citep{abdelhamid2025advisely}.

Several pedagogical frameworks underpin the adoption of LLMs: Bandura's Social Cognitive Theory emphasizes learning through observation \citep{glanz2015health}; Constructivism and Connectivism highlight active and networked learning \citep{vygotsky1978mind, siemens2005connectivism}; and Krashen's SLA theory underscores the importance of comprehensible, low-stress input \citep{krashenESOLCPD}. Yet, concerns remain regarding AI hallucinations, overreliance, and ethical implications \citep{naz2024chatgpt3, xie2024reform, bender2021dangers}. Students may benefit from anonymity and feedback, but also risk disengagement and diminished critical thinking \citep{labadze2023role, carbonel2024chat}.

\subsection{Pedagogical Challenges}
LLMs now generate educational outputs beyond text summaries and Q\&A, including slide decks and podcasts. These formats have the potential to enhance accessibility and engagement \citep{sheridan2020teaching, widodo2019investigating}. However, their rise introduces pedagogical and institutional concerns: reduced human interaction, plagiarism, unreliable detection tools, and algorithmic bias \citep{cotton2023chatting, mitsloan2025aidetectors, abid2021persistent}. Foundational theories by Rogers \citep{rogers1969freedom} and Freire \citep{freire2005pedagogy} emphasize relational and critical learning that may be eroded by overdependence on AI.

To ensure responsible integration, best practices call for institutional policy development, AI literacy promotion, ethical guidelines, and redesigned assessments that preserve critical thinking and originality. These efforts must be grounded in transparency and ongoing collaboration between educators and technologists.

\subsection{Evaluating LLM Outputs}
Robust evaluation is critical to ensure the quality of LLM outputs. Traditional metrics such as BLEU, ROUGE, and METEOR assess token overlap but fail to capture semantic depth \citep{chang2023surveyevaluationlargelanguage}. BLEURT improves on this with contextual sensitivity \citep{evaluationChallenges2024}, though all remain limited for creative, open-ended tasks.

Human evaluation remains the gold standard for fluency, coherence, and relevance, but is time-consuming and subject to inter-rater variability \citep{evaluationChallenges2024}. In response, \textit{LLM-as-a-Judge} frameworks prompt models to evaluate other outputs, offering scalability and task-specific assessments \citep{surveyLLMjudge2024, databricks2023llm}. These methods use pointwise, pairwise, or listwise formats and can be enhanced through few-shot prompting, Chain-of-Thought reasoning, and structured frameworks like G-EVAL \citep{liu2023geval}.

However, LLM evaluation introduce new concerns, such as verbosity and self-preference bias, presentation sensitivity, and vulnerability to adversarial manipulation \citep{panickssery2024llmevaluatorsrecognizefavor, wang2023largelanguagemodelsfair}. Their lack of domain expertise and hallucination tendencies also hinder reliability \citep{chen2024humansllmsjudgestudy}. Meta-evaluation tools and benchmarks like SummEval and Chatbot Arena help align model-based assessments with human judgments \citep{zheng2023judgingllmasajudgemtbenchchatbot}.

\subsection{Future Research Directions}
Despite advances, gaps remain. Future evaluation systems must become modular and domain-adaptive, especially for complex outputs like multimodal educational content. Bias mitigation, validation pipelines, and inclusive evaluation standards are essential. This study contributes to filling this gap by adapting evaluation methodologies to assess non-traditional educational outputs and improve alignment with pedagogical goals.

\section{Methodology}
\subsection{Data Source and Design}
The dataset for this study was built from academic papers retrieved from a public university research archive. A focused subset of ten open-access documents was selected to support the generation of summaries, slides, Q\&A responses, and podcasts. These papers centered on knowledge management to facilitate consistent testing across models. A custom-built web scraper (available in the repository) automates future data expansion by extracting PDFs from the university's portal (1973--2025), but only the selected sandbox corpus was used to ensure output comparability. The documents were cleaned, reformatted, and chunked to enable robust retrieval and generation. This controlled setup provided a realistic yet manageable academic testbed.

\subsection{Model Architectures}
\subsubsection{LLaMA 3.3 - 70B Instruct}
Meta's LLaMA 3.3 70B Instruct was chosen for its long-context handling (up to 128k tokens), multilingual coverage, factuality, and reasoning capabilities \citep{llama32024herd}. Key improvements include curated pre/post-training data, dense Transformer architecture, and supervised fine-tuning. It showed strong performance across benchmarks such as MMLU, GSM8K, and HumanEval. Initial tests with LLaMA 3.3 7B showed reasoning and capacity limitations, justifying the use of the 70B variant.

\subsubsection{GPT-4o-mini}
OpenAI's GPT-4o-mini was used as a second model. Built on distillation, it retains a 128k token context and offers multimodal input and low-latency outputs \citep{openai2024gpt4o}. Its low cost and safety-enhancing instruction hierarchy \citep{wallace2024instructionhierarchytrainingllms} made it ideal for comparative academic deployment. While GPT-4 (at the time of the study, this model was state of art, now it has been replaced by GPT-5) balances performance with cost-efficiency \citep{ong2024gpt4omini}. 

\subsection{Preprocessing Pipeline}
Academic PDFs were parsed using PyMuPDF \citep{PyMuPDF}, cleaned to remove citations, formatting artifacts, and hyphenated line breaks. Text was chunked with LangChain's RecursiveCharacterTextSplitter to create overlapping, semantically coherent segments. Each chunk includes metadata and source attribution. 

\subsection{Retrieval Systems}
Two retrieval pipelines were used:
\begin{itemize}
    \item \textbf{Advanced RAG}: This combines BM25 lexical matching \citep{manning2009information, robertson2009probabilistic} with BERT embeddings \citep{devlin2019bert} and ChromaDB \citep{chromadb}. Named Entity Recognition using spaCy \citep{Abhisarangan2023} refines queries and filters contextually relevant results. Alias-based filtering and scoring adjustments improve document-specific targeting.
    \item \textbf{Graph RAG}: Inspired by Microsoft's GraphRAG \citep{edge2025localglobalgraphrag}, this pipeline integrates a manual knowledge graph constructed with NetworkX and igraph \citep{networkxdocs, igraphsite}. Sentence embeddings from MiniLM-L6-v2 \citep{reimers2019sentencebertsentenceembeddingsusing} drive semantic retrieval. Community detection using the Leiden algorithm \citep{traag2019leiden, Clauset_2004} and cosine similarity \citep{salton1983introduction} structure and prioritize answers.
\end{itemize}

\subsection{Output Generation}
Four types of outputs were generated:
\begin{itemize}
\item \textbf{Q\&A Chatbot}: Answers queries over academic papers using Advanced and Graph RAG.
\item \textbf{Summarization}: Produces structured abstracts from a single document.
\item \textbf{Podcast}: Transforms papers into a scripted dialogue, enhancing engagement.
\item \textbf{Slide Decks}: Creates teaching slides with structured academic bullet points.
\end{itemize}

\subsection{Prompt Engineering}
Prompts were role-based (example role of "academic research assistant") and structured. For Q\&A, LLaMA prompts included behavioral constraints and end-of-response delimiters. GPT-4o-mini prompts emphasized citation grounding. For slides, prompts iteratively generated titles and bullets. Podcasts used dual-character dialogue based on Meta's LLaMA Cookbook \citep{llama_cookbook}. Prompt strategies evolved through testing, mitigating issues like hallucinations and verbosity \citep{roleplay2023}.

\subsection{Evaluation Framework}
The evaluation combined human judgment and LLM-based assessment in a two-phase study. In the first phase, eleven human participants evaluated anonymized chatbot outputs generated from two academic papers using both Advanced and Graph RAG architectures. Evaluations included:
\begin{itemize}
\item \textbf{Graded evaluation}: Participants rated outputs on a 1--5 scale across coherence, correctness, usefulness, and more.
\item \textbf{Pairwise comparison}: Participants selected the preferred output from two options.
\end{itemize}

To minimize subjectivity, scales included standardized criteria, and order randomization was used to counteract positional bias \citep{wang2023largelanguagemodelsfair}. Participants were granted full access to the source documents prior to rating to ensure contextual awareness.

In the second phase, external models (Claude and Deepseek) served as LLM-as-a-Judge evaluators \citep{panickssery2024llmevaluatorsrecognizefavor}. These systems were prompted with the same tasks, instructed to assume the role of critical academic evaluators, and graded outputs using the same criteria. The use of independent models avoided self-preference bias, a known issue in evaluation \citep{surveyLLMjudge2024}.

Evaluation categories were drawn from established research benchmarks \citep{fabbri2021summeval, kocisky2018narrativeqa, liu2023geval}:
\begin{itemize}
\item \textbf{Coherence, Fluency, Relevance, Correctness, Completeness}
\item \textbf{No Hallucinations, Reasoning, Usefulness, Consistency, Engagement, Overall Satisfaction}
\end{itemize}

Importantly, all evaluations were reference-free: no ground-truth answers were provided. Instead, judgments were based on alignment with the academic documents reviewed in advance. This approach reflects real-world applications where responses must be judged for informativeness and coherence, not simple lexical matching.

This multifaceted evaluation design enabled comparative insight into how architecture and model design influence academic output quality, especially in the context of non-traditional educational materials such as podcasts and slides.

\section{Results and Findings}
LLaMA 3.3 70B and GPT-4o-mini were tested with Retrieval-Augmented Generation to generate academic outputs: Q\&A responses, slide decks, and podcast scripts. The objective was to compare performance across these modalities and evaluate the consistency of human versus LLM-based judgment frameworks.

\subsection{Human Evaluation Results}
\subsubsection{Q\&A}
\begin{figure} [!ht]
    \centering
    \includegraphics[width=1\linewidth]{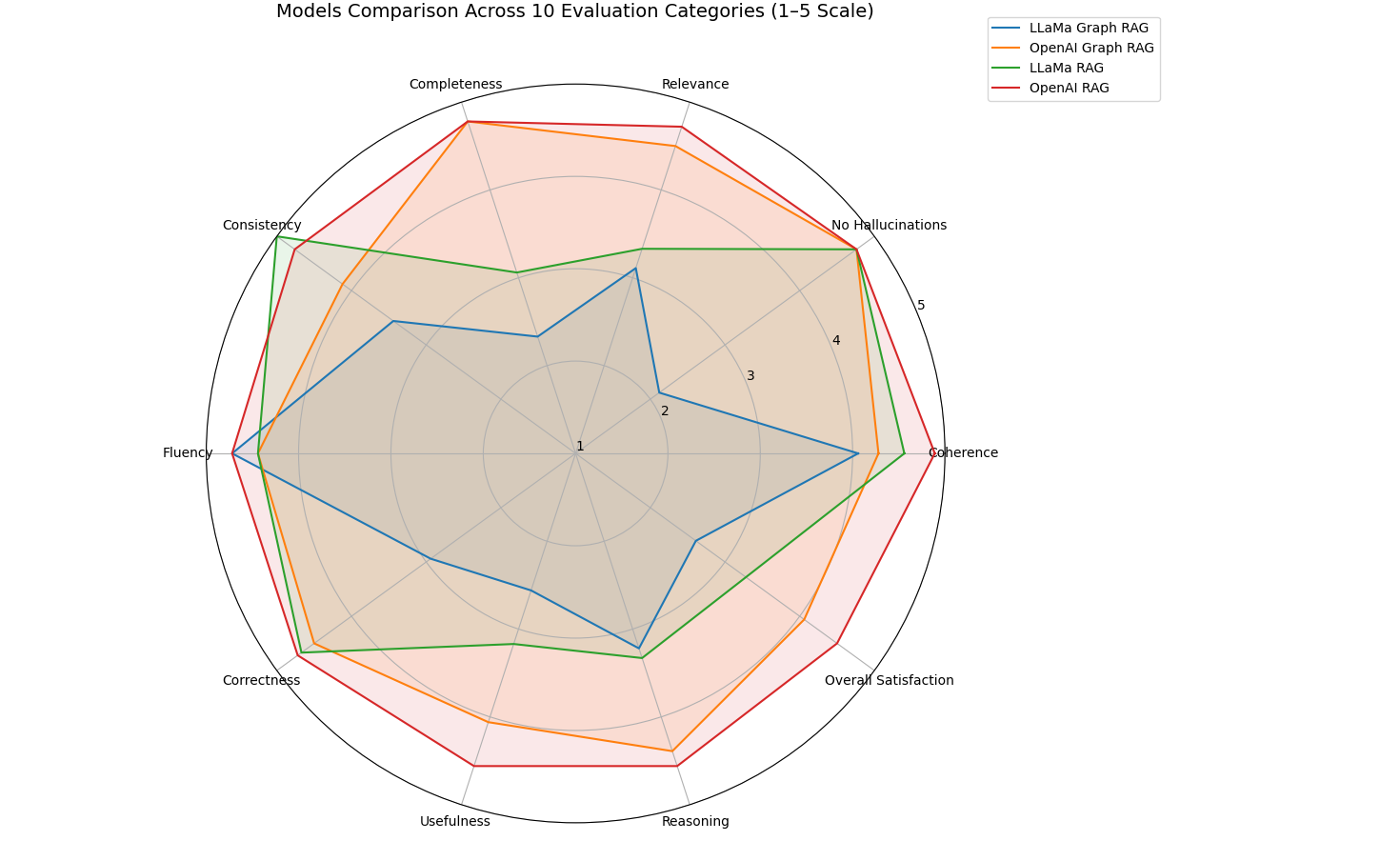}
    \caption{Human Evaluation Grading Score among 10 categories for 2 models and 2 RAG techniques}
    \label{fig:qa_human_grading}
\end{figure}
Figure~\ref{fig:qa_human_grading} presents the average human evaluation scores across ten quality dimensions for four model-architecture combinations. GPT Advanced RAG consistently achieved the highest scores, particularly in usefulness, reasoning, and overall satisfaction, indicating superior performance in academic Q\&A. GPT Graph RAG followed closely, with marginally lower scores. In contrast, LLaMA Graph RAG consistently underperformed, especially in detecting hallucinations, completeness, and usefulness, reflecting a high tendency to introduce inaccurate or incomplete information. All models performed similarly in fluency and coherence, suggesting that even less accurate models can produce well-written text.

The pairwise comparison (Table~\ref{tab:pairwise-summary}) reinforced GPT Advanced RAG's superiority, achieving a 67\% win rate. However, results differed slightly from the graded scores: LLaMA Advanced RAG was preferred over GPT Graph RAG (58\% vs. 27\%), and even LLaMA Graph RAG outperformed GPT Graph RAG in direct comparisons (48\% vs. 27\%). A detailed pairwise comparison reveals that while GPT Advanced RAG dominated all models, its lead over LLaMA Advanced RAG was narrower (56\%), indicating closer competition. These discrepancies suggest that pairwise judgments capture subjective preferences, often influenced by clarity and structure. Participants generally favored concise, well-organized responses, a strength of GPT models.

\begin{table}[ht]
\begin{center}
\begin{tabular}{|l|c|c|}
\hline
\textbf{Model} & \textbf{Wins} & \textbf{Winning \%} \\
\hline
LLaMA Graph & 4.8 & 48\% \\
\hline
GPT Graph & 3.0 & 27\% \\
\hline
LLaMA Advanced & 5.8 & 58\% \\
\hline
GPT Advanced & 7.4 & 67\% \\
\hline
\end{tabular}
\caption{Pairwise comparison summary of Q\&A outputs: Total wins and winning percentage per model based on human evaluations.}
\label{tab:pairwise-summary}
\end{center}
\end{table}

\subsubsection{Non-Traditional Academic Outputs}
\begin{figure}[!ht]
    \centering
    \includegraphics[width=1\linewidth]{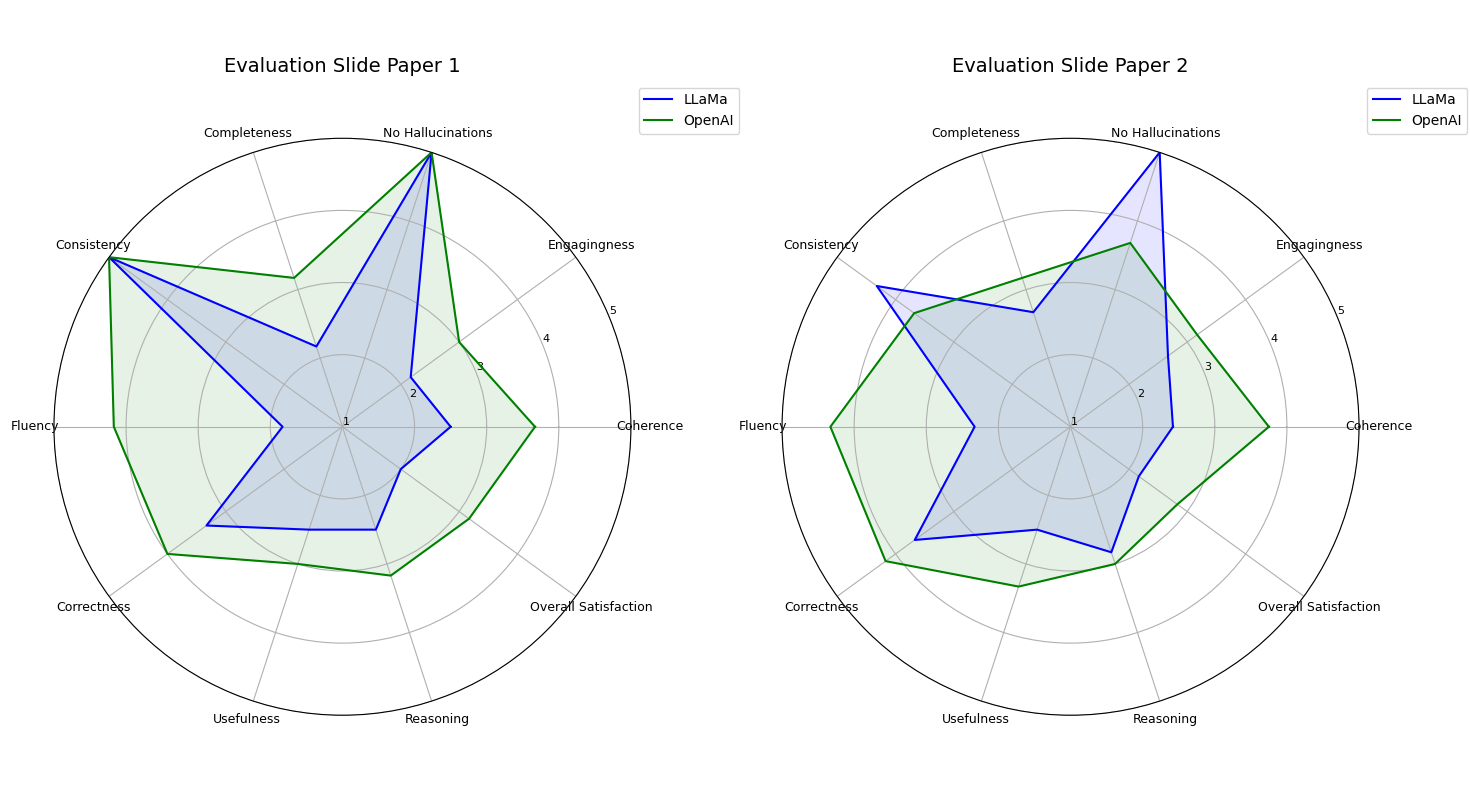}
    \caption{Human-graded evaluation of Slide Deck generation for two academic papers}
    \label{fig:slide-eval}
\end{figure}

\begin{figure}[!ht]
    \centering
    \includegraphics[width=1\linewidth]{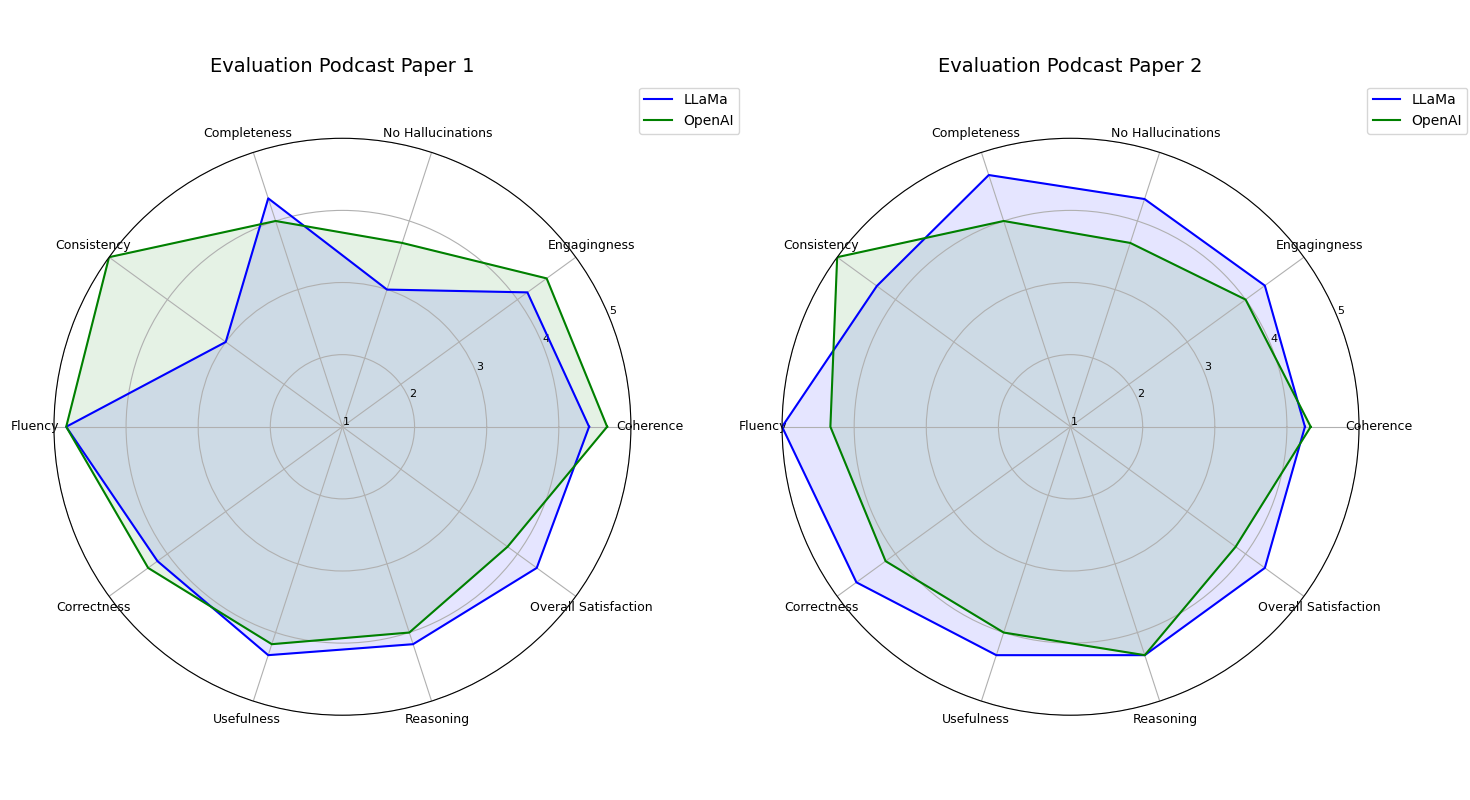}
    \caption{Human-graded evaluation of Podcast Script generation for two academic papers}
    \label{fig:podcast-eval}
\end{figure}

Figures~\ref{fig:slide-eval} and~\ref{fig:podcast-eval} report graded evaluations for slide and podcast tasks. In slide deck generation, GPT models outperformed LLaMA in both papers, particularly in fluency, coherence, and correctness. LLaMA only surpassed GPT in No hallucinations and consistency for Paper 2, indicating better factual fidelity. In podcast generation, GPT again led for Paper 1, excelling in engagingness and coherence. However, LLaMA overtook GPT in Paper 2 across most dimensions, including reasoning and completeness, suggesting higher adaptability to certain texts.

Pairwise preferences (Table~\ref{tab:pairwise-nonstandard}) mirrored these trends: GPT was preferred for slide decks in both papers (100\% and 60\%), while LLaMA was favored for podcasts (60\% in both papers). Feedback indicated GPT slides were more structured and pedagogically effective, whereas LLaMA’s podcasts were perceived as more professional and academically focused. Some raters appreciated GPT’s accessibility and summarization, while others valued LLaMA’s formality. Overall, podcasts were highly appreciated for enhancing understanding, with several evaluators suggesting their integration into study routines.

\begin{table}[!ht]
\centering

\label{tab:pairwise-nonstandard}
\begin{tabular}{|l|c|c|}
\hline
\textbf{Paper} & \textbf{LLaMA Slides} & \textbf{GPT Slides} \\
\hline
Paper 1 & 0\% & 100\% \\
\hline
Paper 2 & 40\% & 60\% \\
\hline
\end{tabular}

\vspace{1em}

\begin{tabular}{|l|c|c|}
\hline
\textbf{Paper} & \textbf{LLaMA Podcast} & \textbf{GPT Podcast} \\
\hline
Paper 1 & 60\% & 40\% \\
\hline
Paper 2 & 60\% & 40\% \\
\hline
\end{tabular}
\caption{Pairwise preferences (\%) for non-traditional academic outputs}
\end{table}

\subsection{LLM-Based Evaluation}
\subsubsection{Q\&A}

\begin{figure}[!ht]
    \centering
    \includegraphics[width=1\linewidth]{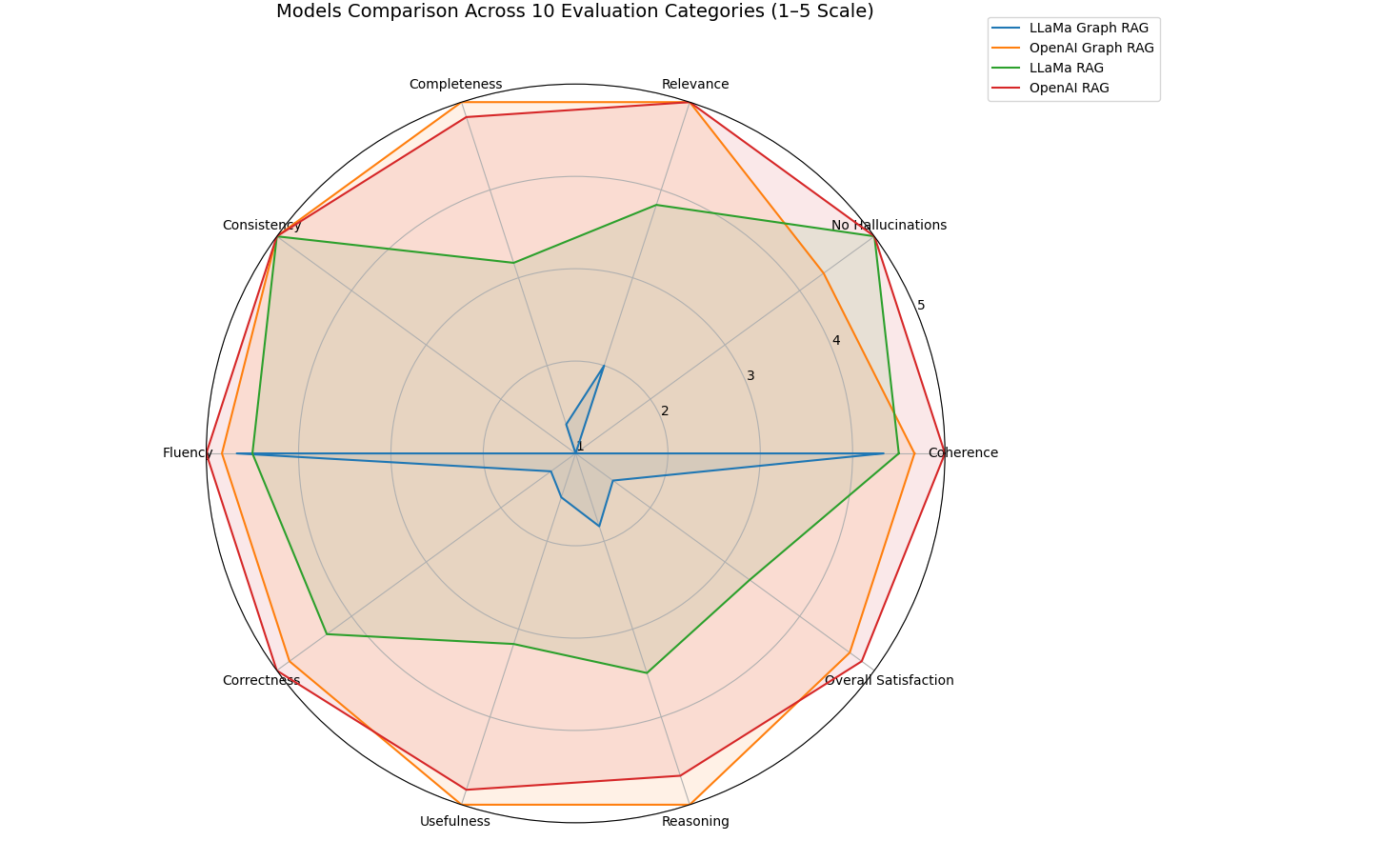}
    \caption{Graded evaluation results from LLM judges (Claude and DeepSeek) across 10 dimensions for Q\&A outputs.}
    \label{fig:qa-llm-avg}
\end{figure}

We evaluated Q\&A outputs using two external LLMs (Claude and DeepSeek), prompted to act as academic judges across the same quality dimensions (Figure~\ref{fig:qa-llm-avg}). The results closely mirror human evaluations: GPT Advanced RAG consistently ranks highest, followed by GPT Graph RAG, LLaMA Advanced RAG, and LLaMA Graph RAG. LLMs were more critical than human evaluators, particularly toward LLaMA Graph RAG, assigning lower scores for correctness, reasoning, and usefulness, suggesting greater sensitivity to factual inconsistencies.

The comparison between Humans and LLMs evaluators shows strong alignment in trends, though LLMs appear stricter on hallucinations and more generous toward elaborate responses. Notably, GPT Graph RAG received higher ratings from LLMs than humans, possibly due to LLMs' preference for verbosity and structured detail. Conversely, GPT Advanced RAG showed near-identical scores between human and LLM judges, reinforcing the validity of its superior performance.

In pairwise comparisons (Table~\ref{tab:llm-pairwise-summary}), GPT Advanced RAG again dominates with an 82\% win rate, followed by LLaMA Advanced RAG (55\%). Unlike the human evaluation, GPT Graph RAG ties with LLaMA Graph RAG (50\% each), indicating LLMs weight completeness and structure more heavily than conciseness or clarity. Detailed breakdowns (Table ~\ref{tab:llm-pairwise-matrix}) confirm GPT Advanced RAG's superiority, winning 75–90\% of head-to-head matchups against all models. Notably, LLMs unanimously preferred LLaMA Advanced RAG over LLaMA Graph RAG, matching graded results.

\begin{table}[!ht]
\begin{center}
\begin{tabular}{|l|r|c|}
\hline
\textbf{Model} & \textbf{Wins} & \textbf{Winning \%} \\
\hline
LLaMA Graph & 5 & 25\% \\
\hline
GPT Graph & 8 & 36\% \\
\hline
LLaMA Advanced & 11 & 55\% \\
\hline
GPT Advanced & 18 & 82\% \\
\hline
\end{tabular}
\caption{LLM pairwise comparison summary: Wins and Winning Percentage per model.}
\label{tab:llm-pairwise-summary}
\end{center}
\end{table}

\begin{table*}[t]
\caption{LLM Pairwise comparison: Percentage of wins for model in the row against model in the column}
\label{tab:llm-pairwise-matrix}
\begin{tabular}{|l|r|r|r|r|}
\hline
\textbf{} & \textbf{LLaMA Graph} & \textbf{GPT Graph} & \textbf{LLaMA Advanced} & \textbf{GPT Advanced} \\
\hline
LLaMA Graph & - & 50\% & 0\% & - \\
GPT Graph & 50\% & - & - & 25\% \\
LLaMA Advanced & 100\% & - & - & 10\% \\
GPT Advanced & - & 75\% & 90\% & - \\
\hline
\end{tabular}
\end{table*}

\begin{figure}[!ht]
    \centering
    \includegraphics[width=1\linewidth]{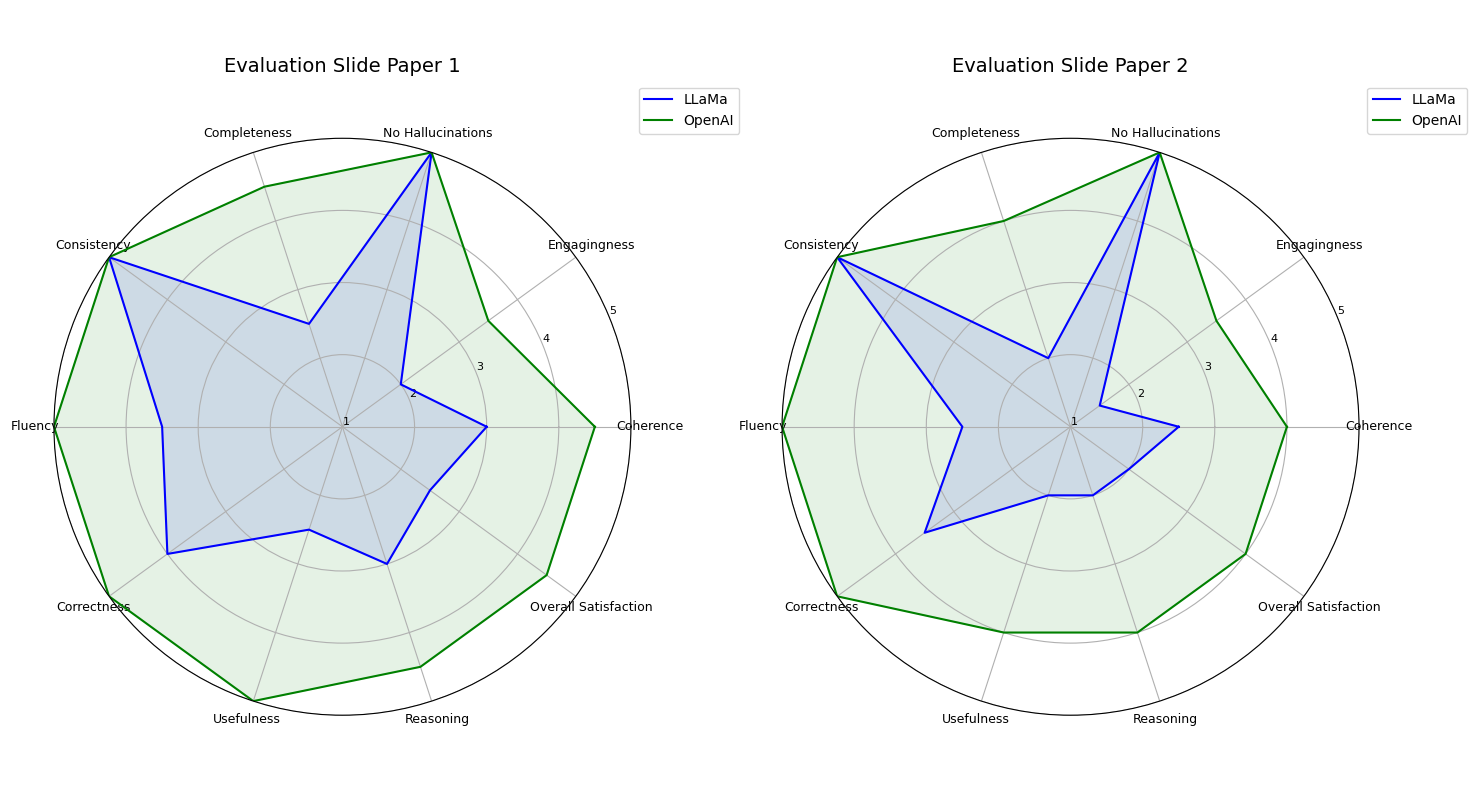}
    \caption{LLM-graded evaluation results for Slide Deck generation across two academic papers.}
    \label{fig:slides-llm}
\end{figure}

\begin{figure}[!ht]
    \centering
    \includegraphics[width=1\linewidth]{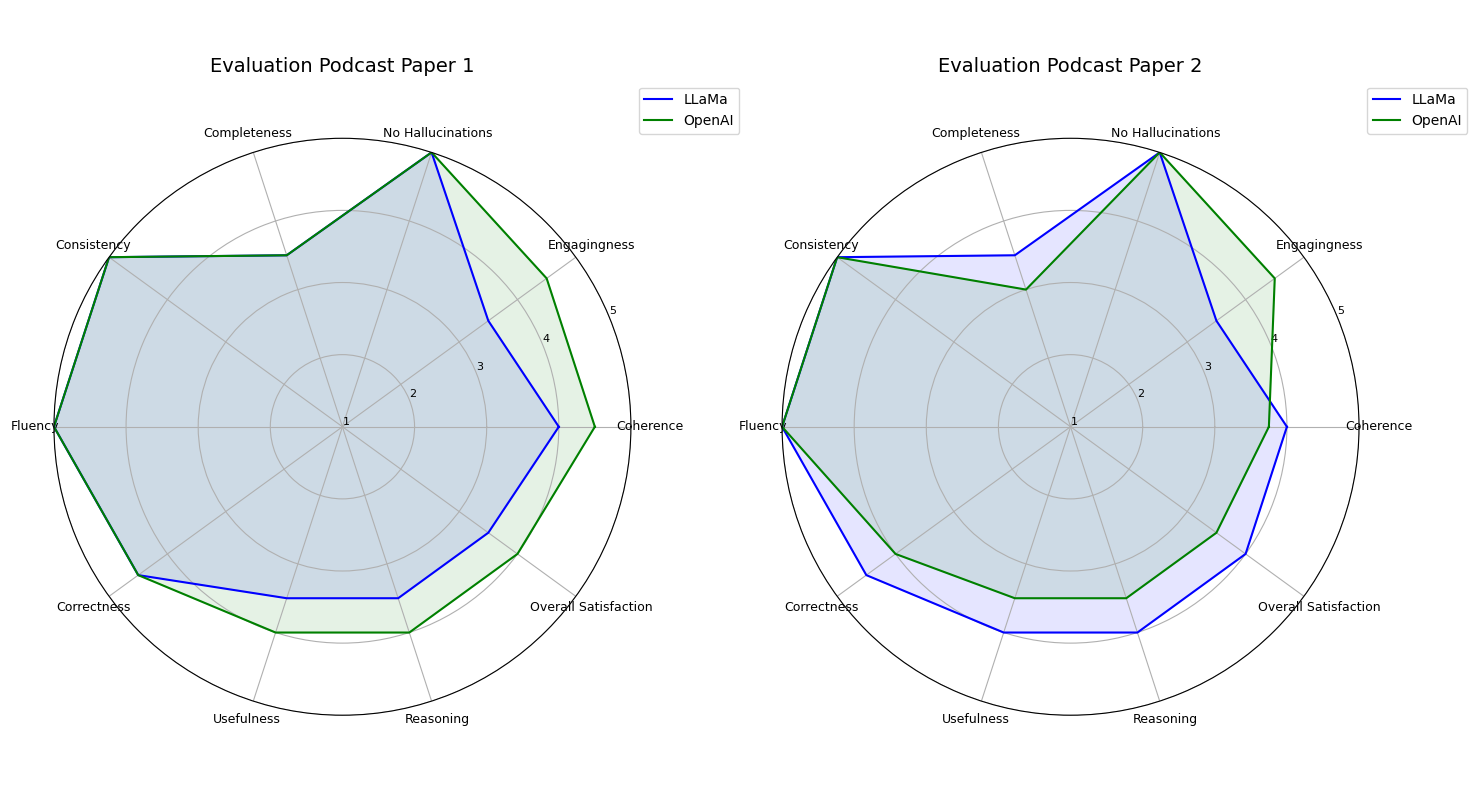}
    \caption{LLM-graded evaluation results for Podcast generation across two academic papers.}
    \label{fig:podcast-llm}
\end{figure}

\subsubsection{Non-Traditional Academic Outputs}
LLMs also scored slide decks and podcasts across ten categories (Figures~\ref{fig:slides-llm} and~\ref{fig:podcast-llm}). For slide decks, GPT models consistently outperformed LLaMA, particularly in coherence, reasoning, and usefulness. Both models scored similarly on consistency and No hallucinations, aligning with human feedback that praised GPT’s structure but acknowledged LLaMA’s factual grounding.

In podcast generation, GPT scored higher for Paper 1, while LLaMA outperformed GPT in Paper 2, especially in completeness and reasoning, suggesting content-specific variability. Human-LLM alignment was generally strong, though LLMs tended to favor GPT more clearly in the slide task, possibly due to an emphasis on structure and linguistic polish.

Pairwise preferences show unanimous LLM support for GPT slides in both papers. Feedback emphasized GPT’s academic tone, completeness, and alignment with research content. For podcasts, preferences were more balanced: LLaMA was preferred for Paper 1, while Paper 2 resulted in a 50–50 tie, with Claude favoring LLaMA’s depth and DeepSeek preferring GPT’s clarity and accessibility. These diverging judgments reflect the trade-off between academic rigor and communicative style.

Overall, LLM-based evaluation confirms human findings: GPT Advanced RAG consistently leads, and GPT models are favored for structured, educational tasks. However, slight divergences, particularly in borderline cases highlight how evaluator type influences quality assessment, with LLMs exhibiting more systematic and format-sensitive criteria.

\subsection{Evaluation Consistency: Standard
Deviation}
To assess evaluator consistency, we analyzed the standard deviation of scores across the QA evaluation categories. Results indicate greater variability among human raters compared to LLMs. For example, GPT Graph RAG had an average standard deviation of 0.72 among humans, versus 0.23 for LLMs; GPT Advanced RAG dropped from 0.51 (human) to 0.17 (LLMs). This suggests that LLMs provide more stable and repeatable evaluations, while human judgments are more subjective, particularly in dimensions like Relevance, Completeness, and Reasoning, notably for LLaMA Advanced RAG.

In slide deck evaluations, human raters again showed higher variance, particularly in Engagingness and Completeness (often >1.0). LLMs, on the other hand, showed minimal disagreement, especially for GPT's slides, where some categories had near-identical scores across both papers.

For podcasts, human evaluations showed moderate variation, with highest deviations in Correctness, Usefulness, and Reasoning (notably for LLaMA Paper 1 and GPT Paper 2). LLM scores exhibited almost no variation (often 0.00), with slight divergence only in LLaMA’s Paper 2 podcast (Coherence and Correctness).

Overall, LLMs demonstrate high inter-rater agreement due to their uniform scoring logic, while human raters provide richer but more variable assessments, particularly valuable for evaluating subjective elements such as tone, delivery, and engagement in creative formats like slides and podcasts. This contrast highlights the trade-off between LLM consistency and human nuance in evaluation.

\subsection{Key Findings}
The evaluation reveals a clear performance hierarchy across all tasks, with GPT Advanced RAG emerging as the most effective model in both human and LLM-based assessments. It consistently scored highest in usefulness, reasoning, and overall satisfaction, and was preferred in pairwise comparisons for Q\&A, slide decks, and podcasts. LLaMA Graph RAG, in contrast, ranked lowest across all dimensions, particularly struggling with correctness, completeness, and hallucination control, indicating limitations in its integration with graph-based retrieval methods.

Human and LLM evaluators showed strong alignment in their rankings, especially for top and bottom performers, though LLMs tended to be more stringent, particularly in detecting hallucinations and inconsistencies. GPT’s outputs were consistently favored in slide deck generation for their structured, pedagogically aligned content, while podcast results were more nuanced. In some cases, especially in Paper 2, LLaMA’s podcasts were preferred for their academic tone and fidelity to the source, whereas GPT’s were recognized for clarity and accessibility, pointing to a trade-off between rigor and engagement.

A notable difference emerged in the consistency of evaluations: LLMs demonstrated far lower standard deviations across all tasks compared to humans, reflecting higher inter-rater agreement and systematic scoring behavior. Human evaluators showed greater variation, particularly in subjective dimensions such as completeness, usefulness, and engagingness, underscoring the diversity of human judgment and sensitivity to stylistic nuances. These findings suggest that while LLMs can replicate human preferences in well-performing outputs, human feedback remains essential for capturing qualitative aspects, especially in creative or presentation-based academic content.

\section{Conclusions}
In addressing our primary research question, we find, based on both quantitative evaluation and qualitative feedback from human raters, that Large Language Models, when combined with Retrieval-
Augmented Generation, can generate high-quality non-traditional academic outputs such as slide
decks and podcasts that are usable and pedagogically valuable. Several evaluators expressed interest in
using these tools to support their own learning, noting that the generated content helped them better
understand the academic material.

Regarding Q1, GPT-4o-mini coupled with our Advanced RAG pipeline was consistently the strongest
generator for traditional and semi-structured tasks: it won 67\% of human pairwise comparisons in the
Q\&A study and an even higher 82\% of pairwise votes from LLM judges, while LLaMA 3.3 70B in its best
configuration (Advanced RAG) reached only 58\% and 55\%, respectively. The same pattern held for slide
decks, while podcasts, however, exposed complementary strengths: humans chose the more academically
dense LLaMA scripts in 60\% of cases for both papers, whereas Claude and DeepSeek split their vote on
Paper 2, signaling that narrative style and audience fit still differentiate the models.

For Q2, the dual-track evaluation framework proved both feasible and informative: LLM-based scores
reproduced the human quality ranking almost perfectly while exhibiting far tighter agreement (average $\sigma = 0.23$ for GPT Graph RAG vs. 0.72 for human graders, and $\sigma = 0.17$ vs. 0.51 on GPT Advanced RAG), showing that automated judges can supply low-variance, reference-free feedback. Yet the larger
human standard deviations (up to 1.57 in “Relevance”) also revealed nuances, such as layout glitches or
tone preferences, that LLMs either overlooked or weighted differently.

Taken together, the results show that (i) state-of-the-art proprietary LLMs, like GPT-4o-mini, combined
with retrieval deliver the highest factual and pedagogical quality across modalities, while open-weight
models, such as LLaMA 3.3 70B, can still excel in conversational, narrative outputs, and (ii) a mixed human-
and-LLM evaluation protocol captures robust quantitative feedback together with qualitative judgments,
such as perceived engagement, tonal appropriateness, and overall resonance, needed for academic adoption.

What this research aims to achieve is just to set one more stepping stone into a better production and understanding of possible generated outputs from the implementation of LLMs as tools in the field of education. Many possible directions were presented in this work, and future research can focus on a multitude of analyses to expand on what has been done so far, such as developing a comprehensive evaluation framework designed specifically for non-traditional education outputs, using benchmarks built on evaluation collected from domain experts. Another future research path could be longitudinal testing in classroom settings to provide stronger evidence of pedagogical value and practical usability. Also, future work should focus on developing or integrating scalable entity and relationship extraction systems that are optimized for domain-specific academic content, and since both Advanced RAG and Graph RAG have their unique advantages that should be leveraged to improve overall performance\citep{han2024raggraphrag}, therefore future research should focus on a possible novel approach to combine the two systems for both effectiveness and efficiency, Lastly, still with the objective of closing some of the gaps currently present, as LLMs increasingly support content creation in sensitive educational contexts, future work should incorporate fairness audits and inclusivity checks into the evaluation pipeline. 

Beyond technical performance, this work underscores a fundamental principle: educational AI, and AI in general, must ultimately serve human needs. Automated evaluation offers scalability, but human judgment remains essential for assessing clarity, usefulness, and pedagogical value. Future research should continue to refine pipelines and metrics, but always with the end user in focus, because the success of these systems will be measured not only by accuracy, but by their ability to support meaningful learning.

\section{Limitations and Ethical Considerations}
In this study, we manually curated entities and relationships for GraphRAG to isolate the impact of knowledge graph structures on QA performance. While this increases best case performance, it does not reflect real-world use, where automated entity extraction is essential. We transparently acknowledge this trade-off, following \citep{ji2021survey}, who recommend human oversight during KG construction to mitigate noise. Our aim was to assess the graph structure's value independently of upstream noise. Evaluation constraints also shaped our findings. Limited evaluators and outputs restrict statistical generalizability and introduce risk of overfitting conclusions. Human evaluation, although essential for nuanced judgments, is inherently subjective. Raters differed in their preferences, some prioritizing clarity, others formal tone or depth, reflected in high standard deviations across scores.
Additionally, hallucination detection varied: LLM judges like Claude were stricter in penalizing unsupported claims, while human evaluators often overlooked subtle hallucinations if the text seemed plausible. Conversely, LLMs showed bias toward stylistically familiar formats and verbosity, possibly favoring outputs that align with their own generation patterns rather than those most pedagogically useful.
Lastly, our reference-free evaluation allowed for flexible judgment grounded in source documents but introduced challenges in quantifying factual precision. Despite these challenges, combining human and LLM assessments yielded a richer understanding of output quality.

The integration of AI chatbots in academia presents both opportunities and risks. From an operational standpoint, AI can streamline lecture preparation, content summarization, slide generation, and academic Q\&A, reducing faculty workload. As demonstrated in our case study, this supports institutional goals of digital transformation and teaching personalization. Beyond education, non-traditional outputs like AI-generated podcasts or slides can support marketing and sponsorship efforts, showcasing innovative teaching strategies. However, the benefits come with infrastructure costs and ethical responsibilities. Institutions must invest in robust systems and oversight to mitigate hallucinations, bias, or inaccuracies, particularly if underfunded or lacking technical capacity.

Societally, AI reshapes education by enabling personalized, 24/7 learning support. Students must now develop AI literacy, critically assess generated content, and understand how to engage with these systems. Teachers, meanwhile, may shift into facilitative roles, emphasizing reflection and ethical reasoning.

However, risks arise around data privacy, especially when commercial APIs like OpenAI are used. Transparency and GDPR compliance are critical. Moreover, algorithmic surveillance may pressure students toward system-optimized behavior, reducing academic freedom. As Williamson (2019) warns, outcome-driven AI risks narrowing pedagogy to what’s measurable.

Bias amplification is another key concern. LLMs may reproduce stereotypes present in training data, harming marginalized groups and perpetuating inequality in knowledge access. Academic integrity is also at stake: distinguishing between student-generated and AI-assisted work grows increasingly difficult, especially as AI detectors remain unreliable.

Lastly, the digital divide risks leaving underfunded institutions behind, widening gaps in access to modern learning tools. Without proactive design and inclusivity measures, academic AI may reinforce, not resolve, existing educational inequalities.

\section{Data and Code}
Because the paper is currently being reviewed for a conference with an anonymous submission requirement, all data and experimental code are provided in an anonymous GitHub repository: \href{https://anonymous.4open.science/r/Chatbot-in-academia-04BE/README.md}{https://anonymous.4open.science/r/Chatbot-in-academia-04BE/README.md}.
So for all resources, consult the link. 

\section{Acknowledgment}
We thank Copenhagen Business School for supporting this research. We are especially grateful to our supervisor and co-author, Professor Daniel Hardt, for his guidance, feedback, and invaluable contributions throughout this project.

\section{Bibliographical References}\label{sec:reference}

\bibliographystyle{format}
\bibliography{languageresource}

\end{document}